\documentclass[Journal]{IEEEtran}

\usepackage{amssymb,color,amsmath}
\usepackage{paralist,cite}
\usepackage{graphicx}
\usepackage{algorithm2e}
\usepackage{algorithmic}

\usepackage{epstopdf}

\newcommand{\nix}[1]{}

\begin{document}
\title{\huge{A Hajj And Umrah Location Classification System For Video Crowded Scenes}}

\author{
{Hossam M. Zawbaa${^\dag}$~~~~~~~~~~~Salah A. Aly$^{\dag\ddag}$}~~~~~~~~~~~Adnan A. Gutub${^\dag}$\\
\medskip
{$^{\dag}$Center of Research Excellence in Hajj and Umrah, Umm Al-Qura University, Makkah, KSA\\ $^{\ddag}$College  of Computers and Information Systems, Umm Al-Qura University, Makkah, KSA}}
\maketitle

\begin{abstract}

In this paper, a new automatic system for classifying ritual locations in diverse Hajj and Umrah video scenes is investigated.  This challenging subject has mostly been ignored in the past due to several problems one of which is the lack of realistic annotated video datasets.  HUER Dataset is defined to model six different Hajj and Umrah ritual locations~\cite{huer2012}.

The proposed Hajj and Umrah ritual  location classifying system  consists of four main phases: Preprocessing, segmentation, feature extraction, and location classification phases. The shot boundary detection and background/foregroud segmentation algorithms are applied to prepare the input video scenes into the KNN, ANN, and SVM classifiers.  The system improves the state of art results on Hajj and Umrah location classifications, and successfully recognizes the six Hajj rituals with more than $\%90$ accuracy. The various demonstrated experiments  show the promising results.\footnote{\noindent ------------------------------------------------------------------------------------- \goodbreak \noindent  Thanks to HajjCoRE, Center of Research Excellence in Hajj~ and~ Umrah at~ UQU,  for supporting this work.\\ Contact: salahaly@uqu.edu.sa}
\end{abstract}

\section{Introduction}\label{sec:intro}
During the last two decades, the field of visual recognition had an outstanding evolution from classifying instances of toy objects towards recognizing the classes of objects and scenes in natural images. Much of this progress has been sparked
by the creation of realistic image datasets as well as by the new, robust methods for image description and classification. We take inspiration from this progress and aim to transfer previous experience to the domain of video recognition
and the recognition of human actions in particular for Hajj and Umrah Videos~\cite{Laptev2008}.

\goodbreak

Action recognition from video shares common problems with object recognition in static images. Both tasks have to deal with significant intra-class variations, background clutter and occlusions. In the context of object recognition
in static images, these problems are surprisingly well handled by a bag-of-features representation \cite{Willamowski2004} combined with state-of-the-art machine learning techniques like support vector machines. It remains, however, an open question whether and how these results generalize to the recognition of realistic human actions, e.g., in feature films or personal videos.

\medskip

Building on the recent experience with image classification, we employ spatio-temporal features and generalize spatial pyramids to spatio-temporal domain. This allows us to extend the spatio-temporal bag-of-features representation with weak geometry. We validate our approach on a standard benchmark \cite{Laptev2004} and show that it outperforms the state-of-the-art. We next turn to the problem of action classification in realistic Hajj and Umrah videos and show promising results for eight very challenging action classes including walking, drinking Zamzam water, sleeping, smiling, eating, praying, sitting, shaving hair, doing ablution, reading the Holy Quran and making duaa. Finally, we present and evaluate a fully automatic setup with action learning and classification obtained for an automatically labeled training datasets~\cite{huer2012}.

In~\cite{huer2012}, the authors proposed new  event recognition  datasets in Hajj and Umrah Videos. They presented two different classes of datasets: \begin{inparaenum}
\item Rituals locations dataset.
\item Human event recognitions dataset.
\end{inparaenum}

This paper is organized as follows.   Section~\ref{sec:systemmodel} presents the system description and proposed new model. In Sections~\ref{sec:preprocessing},~\ref{sec:eventrecognition},~\ref{sec:classifications}, we developed two algorithms for video pre-processing and background/foreground segmentations. Simulation results are presented in Section~\ref{sec:analysis}. Section~\ref{sec:previouswork} presents the related work and background.
Finally, the paper is concluded in Section~\ref{sec:conclusion}.

\begin{figure}[t]
\begin{center}
\includegraphics[width=8.5cm, height=8cm]{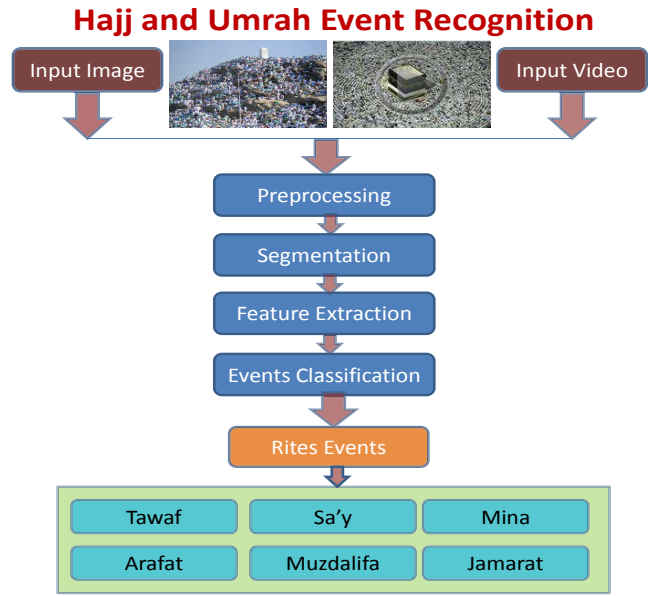}
\end{center}
\caption{The System Model consists of four main phases: Pre-processing, segmentation, feature extraction, and classifying location.}
\label{fig:SystemModel}
\end{figure}

\section{System Model and Description}\label{sec:systemmodel}

In this section we define our research problem and propose a new solution model.

\begin{figure}[h]
\begin{center}$
\begin{array}{cc}
\includegraphics[width=8.5cm, height=8cm]{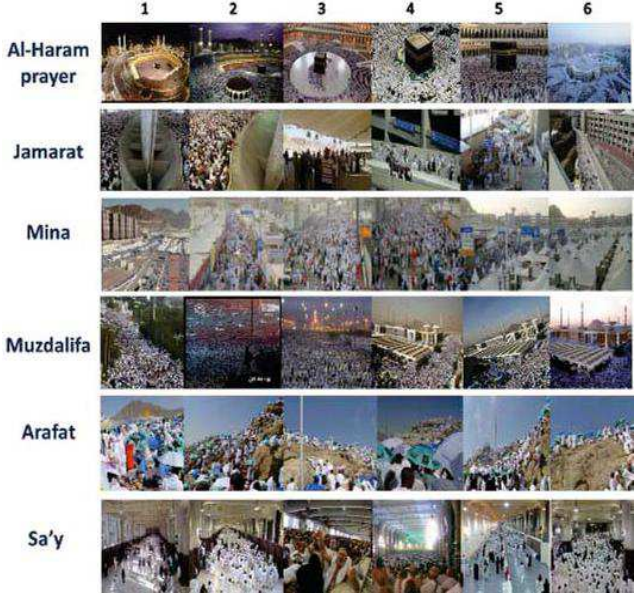}
\end{array}$
\end{center}
\caption{Hajj and Umrah Pilgrim Events Recognition Datasets. Various images are taken from different places representing Hajj and Umrah rituals.}
\label{fig:Dataset2}
\end{figure}

\bigskip
\noindent\textbf{Problem Definition:}  The problem can be defined as how to detect the ritual locations (Tawaf, Sa'y, Arafat, Muzdalifa, Mina, Jamarat). In particular, the proposed framework is capable of recognizing a wide range of location classification in diverse Hajj and Umrah video and image scenes under different conditions. The main goal is to develop the proposed system to detect the ritual locations, see Fig.~\ref{fig:Dataset2}.

\bigskip

\noindent\textbf{Proposed System Model: }
This paper presents a location classification system during Hajj and Umrah seasons in the two Holy cities of Makkah and Madina. The proposed system is composed of four phases as follows:

\begin{enumerate}
 \item {\emph{Pre-processing phase}} that segments the whole video stream into small video scenes and capturing the candidate key frames from it.
  \item {\emph{Segmentation phase}} that separates the foreground objects from the background from Hajj and Umrah images and videos.
   \item {\emph{Feature extraction phase}} that defines the interest points and its description for the background and foreground images.
   \item {\emph{classification phase}} that applies two types of classification, the pilgrim events for the foreground features, and the rites events for the background features resulted from the feature extraction phase.
\end{enumerate}

These four main phases are described in detail in the next four sections along with the characteristics feature steps involved in each phase.\\

\section{Pre-processing Phase}\label{sec:preprocessing}

The goal of the pre-processing phase is to segment the video stream into video shots and select the k-frames. The process of detecting the actual boundary of shots depends on object and camera motion by frame correlation. Each shot is defined as a sequence of frames captured by a single camera in a single continuous action in time and space \cite{Song2009}. The correlation between frames in the same shot is a very important indicator to detect the similarity between them and when there is noticeable change we can conclude the starting of new shot as previously mentioned.

There are two types of transitions depending on the camera movement: (1) instant (cut) transition and (2) gradual transition. When there is a special editing effect during the movie, it's called a gradual
transition, whereas the instant transition has no editing effects and  is more accurate than the gradual one.
Most of Hajj and Umrah videos have a gradual transition, because these videos have been recorded using digital cameras, mobile phones by amateur individuals who are not professional or by a fixed El-Harram camera which needs a permission from Saudi authorities. The shot boundary detection algorithm (SB-Alg)~\ref{lst:Shotboundary} is adapted in order to generate video shots. \\

\begin{algorithm}[t!]
\KwIn{The input video streams}%
\KwOut{Small video shots} 

\ForEach{Input video streams}%
  {
    - Capture frames from the input video.\\
    - Convert the gradual transition into cut transition by using frame skipping k (k = 10).\\
    \ForEach{frame}
    {
        - Convert frames from $RGB$ to $HSI$ color space using Equations ~\eqref{Eq.104}, ~\eqref{Eq.105}, and ~\eqref{Eq.106}.\\
        - Calculate the motion difference between current hue frame and the next hue k-frame.\\
        - Divide the original color frames into $64 \times 64$ block size.\\
        - Calculate the percentage of changed blocks between the current frame and the next k-frame.\\

          \If{(motion difference $> Thr$) and (block changing percentage $> 0.25$)}%
                {
                 - Mark a new shot.\\
                 - Select the k-frames using $k = 10$ frames.%
                }%
    }
 }%
\mbox{}\\
\caption{A shot boundary detection Algorithm (SBD-Alg) for Hajj and Umrah videos} \label{lst:Shotboundary}
\end{algorithm}

The following equations describe the proposed algorithm. These equations convert frames from $RGB$ to $HSI$ color space \cite{Gonzalez1993}.
    \begin{equation}\label{Eq.104}
        H = cos^{-1} \frac{(0.5 * (R-G) + (R-B))}{\sqrt{((R-G)^{2}+(R-B)(G-B)}},
        \end{equation}

        \begin{equation}\label{Eq.105}
          S = 1 - \frac{3}{R+G+B} * \min(R,G,B),
        \end{equation}

        \begin{equation}\label{Eq.106}
          I = \frac{1}{3}(R + G + B),
        \end{equation}
where H, S and I are Hue, Saturation, Intensity, respectively; and R, G, and B are the traditional Red, Green and Blue colors.

\section{Segmentation Phase}\label{sec:Pre-processing}

Separating foreground objects from natural images and videos play an important role in image and video editing tasks. Despite extensive study in the last two decades, this problem still remains challenging. Segmenting spatio-temporal video objects from a video sequence are even harder since extracted foregrounds on adjacent frames must be both spatially and temporally coherent. This section demonstrates efficient foreground extraction methods and systems by combining advanced computational algorithms \cite{Gonzalez1993}.

\subsection{Background Foreground Model}

The first step of background subtraction is to setup the background model or the reference image. The background is modeled in two distinct parts: a luminance model and a color model. Input video streams or input images have three channels with RGB components, but they are very sensitive to noise and changes of lighting conditions. Therefore, we use a luminance component of  color images for initial object segmentation. Image luminance $Y$ is calculated with the following Equation~\cite{Shapiro2001}:

\begin{equation}\label{Eq.100}
   Y = 0.299 × R + 0.587 × G + 0.114 × B.
\end{equation}

However, the luminance component changes drastically by shadows of objects in the background regions and the reflection of lighting in the foreground regions. The HSI color space (Hue, Saturation, Intensity) is often used, because it corresponds better to how people experience color than the RGB color space does. As Hue varies from $0$ to $1.0$, the corresponding colors vary from red through yellow, green, cyan, blue, magenta, and back to red. Therefore, there are actually red values both at $0$ and $1.0$. As saturation varies from $0$ to $1.0$, the corresponding colors (Hues) vary from unsaturated (shades of gray) to fully saturated (no white component). As value, or brightness, varies from $0$ to $1.0$, the corresponding colors become increasingly brighter. Equations~\eqref{Eq.104},~\eqref{Eq.105}, and~\eqref{Eq.106} illustrate the HSI color space \cite{Gonzalez1993}.

\subsection{Foreground Segmentation}

The background subtraction is based on the varying features of H, S, and V for each pixel in the background. After that subtract the current frame from the background model or reference image to obtain the moving video objects or the foreground image. So, to achieve this purpose, we use Gaussian normal distribution as a statistical model to model the distribution of each color component of the background's pixels. So, the background model is to acquire the mean and standard deviation of the color components of the background's pixels. By detecting the variation of the pixels in the background model, the video objects can be segmented. Now the background subtraction algorithm~\ref{lst:FgSegmentation} is summarized as follows~\cite{Shapiro2001}.


\begin{algorithm}[t!]
\KwIn{Video stream}%
\KwOut{Background and Foreground images sequence from the input video stream} 

\ForEach{Input video stream}%
  {
    Capture frames from the input video

    \ForEach{frame}
    {
        Convert frames from $RGB$ to $HSI$ color space using Equations ~\eqref{Eq.104}, ~\eqref{Eq.105}, and ~\eqref{Eq.106} \cite{Gonzalez1993}.
    }

\ForEach{(H, S, and I) color components in HSI color space}
        {
        Computing the $\mu$ and $\sigma$ parameters for the background statistical model for all pixels from the first $N$ frames.
        }

 \ForEach{frame}
    {
    \ForEach{pixel}
        {
            Calculate the background pixels distribution $P(x)$ using gaussian normal distribution as shown in Equation ~\eqref{Eq.107}:
        }
        Calculate the correlation distance measure between the current frame and the background model or reference image as shown in Equations ~\eqref{Eq.108}:\\
           \If {$Corr(x,y) < Thr$}
                 {$x$ belongs to the background pixels, Otherwise it belongs to the foreground pixels.}
     }
        Update the background statistical model parameters $\mu$ and $\sigma$.
}%
\mbox{}\\
\caption{Video Background and Foreground segmentation} \label{lst:FgSegmentation}
\end{algorithm}

     The following equation describes and calculates the background pixels distribution.

         \begin{equation}\label{Eq.107}
            P(x) = \frac{1}{\sqrt{2 \Pi}\sigma}exp^{\frac{-(x-\mu)^{2}}{\sigma^{2}}},
            \end{equation}

            where $P(x)$ is the probability distribution function, $\mu$ is the mean, and $\sigma$ is the standard deviation.\\

   The correlation distance measure is described in the following equation.

             \begin{equation}\label{Eq.108}
            Corr (x, y) = 1- \frac{(x - \widetilde{X}).(y - \widetilde{Y})}{\|(x - \widetilde{X})\|  \|(y - \widetilde{Y})\|},
         \end{equation}
         where $\widetilde{X}$ is the mean of the current image, and $\widetilde{Y}$ is the mean of the reference image.\\

\section{Feature Extraction Phase}\label{sec:eventrecognition}

In this section, we split feature extraction into two stages:
\begin{enumerate}
 \item Feature detectors: The resulting features will be subsets of the image domain, often in the form of isolated points, continuous curves or connected regions. Feature detection is how to find some interesting points (features) in the image, as example (find a corner, find a template, and etc).

 \item Feature descriptors: It represents the interesting points we found to compare them with other interesting features in the image, for example (the local area intensity of this point, the local orientation of the area around the point, and etc).
\end{enumerate}

\medskip
Most feature detectors involve the computation of derivatives or more complex measures such as the second moment matrix for the Harris detector or entropy for the salient regions detector. Since this step needs to be repeated for each and every location in feature coordinate space which includes position, scale and shape. This makes the feature extraction process computationally expensive, which is
not suitable for many applications.

We describe several feature detectors that have been developed with computational efficiency as one of the main objectives. The SIFT uses the Difference of Gaussian (DoG) detector approximates the Laplacian uses multiple scale space pyramids, in addition SURF  uses  integral images to efficiently compute a rough approximation of the Hessian matrix. FAST evaluates only a limited number of individual pixel intensities using decision trees.

The proposed system will model the Hajj and Umarh events through the information obtained with the tracking of the feature points. We rely on those motion vectors of feature points computed over multiple frames. After the pre-processing phase, we have two separates sequences of video frames, background and foreground. In the proposed system, we use the modified scale-invariant feature transform (SIFT) algorithm identifies features of an image that are distinct. These features can be  used to identify similar or identical objects in other images  as shown in Algorithm~(\ref{lst:SIFT}).

\subsection{Scale-Invariant Feature Transform (SIFT)}

SIFT consists of four major stages: scale-space extrema detection, keypoint localization, orientation assignment, and keypoint descriptor. The first stage used difference-of-Gaussian function to identify potential interest points, which were invariant to scale and orientation. DOG was used instead of Gaussian to improve the computation speed. Gaussian image pyramid $L(x, y, \sigma)$ is generated by successively filtering the image $I (x, y)$ with Gaussian filter $G(x, y, \sigma)$ according to Equation ~\eqref{Eq.109}, and ~\eqref{Eq.110}. Adjacent Gaussian images are subtracted to produce the difference-of-Gaussian images (DoG) as in Equation ~\eqref{Eq.111}, which approximate the Laplacian of a Gaussian filtering \cite{Lowe2004}.\\

In the keypoint localization step, for more accurate localization of the keypoint and to remove the keypoints with low contrast, we just discard keypoints in
which the absolute value of DoG pyramid at the interval they are detected is smaller than certain threshold. The keypoint is not lying on a strong edge. For this reason, we use the discrete differences between neighboring pixels around the keypoints to calculate the Hessian matrix. The interest points detected with the determinant of the Hessian to compute the principal curvatures and eliminate the keypoints that have a ratio between the principal curvatures lower than $Thr$.\\

An orientation histogram is formed from the gradient orientations of sample points within a region around the keypoint. Each sample added to the histogram is weighted by its gradient magnitude and by a Gaussian-weighted circular window. We only assign one orientation to each keypoint which corresponds to the peak of the histogram. According to the experiments, the best results were achieved by 4 x 4 array of histograms with 8 orientation bins in each. So the descriptor of SIFT that was used is $4 x 4 x 8 = 128$ dimension vector~\cite{Lowe2004}.
After that we applying sparse coding based on SIFT features. The sparse coding is to represent input vectors approximately as a weighted linear combination of the basis vectors which capture from the high level patterns in the input data as shown in Algorithm~(\ref{lst:SIFT})~\cite{Lowe2004}.

\begin{algorithm}[t!]
\KwIn{The background or foreground images sequence}%
\KwOut{Features and labels for each new locations} 

\ForEach{background image}%
{
    - Build a Gaussian image pyramid L(x, y, $\sigma$) using Equations~\eqref{Eq.109},~\eqref{Eq.110}, and~\eqref{Eq.111}.

    - Calculate the Hessian matrix as in Equation ~\eqref{Eq.112}.

    - Calculate the determinant of the Hessian matrix as in Equation~\eqref{Eq.113} and eliminate the weak keypoints.

    - Calculate the gradient magnitude and orientation as in Equations~\eqref{Eq.114} and~\eqref{Eq.115}.

    - Apply the sparse coding feature based on SIFT descriptors as in Equations~\eqref{Eq.115-1} and~\eqref{Eq.115-2}.
}%
\caption{Feature extraction using SIFT} \label{lst:SIFT}
\end{algorithm}

 \begin{equation}\label{Eq.109}
            G(x, y, \sigma) = \frac{1}{2 \Pi \sigma^{2}} exp^{\frac{-(x^{2} + y^{2})}{2 \sigma^{2}}},
    \end{equation}

    \begin{equation}\label{Eq.110}
            L(x, y, \sigma) = G(x, y, \sigma) * I(x,y),
    \end{equation}

    \begin{equation}\label{Eq.111}
            D(x, y, \sigma) = L(x, y, k \sigma) - L(x, y, \sigma),
    \end{equation}

    where $\sigma$ is the scale parameter, G(x, y, $\sigma$) is Gaussian filter, I(x, y) is smoothing filter, L(x, y, $\sigma$) is Gaussian pyramid, and D(x, y, $\sigma$) is difference of Gaussian (DoG) \cite{Wu2010}.

\begin{equation}\label{Eq.112}
H =
 \begin{bmatrix}
  I_{xx}(x,\sigma) & I_{xy}(x,\sigma) \\ \\
  I_{xy}(x,\sigma) & I_{yy}(x,\sigma)
 \end{bmatrix}.
\end{equation}

Where $I_{xx}$ is the second-order Gaussian smoothed image derivatives which detect signal changes in two orthogonal directions.

\begin{equation}\label{Eq.113}
    Det(H) = I_{xx}(x,\sigma) I_{yy}(x,\sigma) - (I_{xy}(x,\sigma))^2
\end{equation}

\begin{eqnarray}\label{Eq.114}
    Mag(x,y) &=&  ((I(x+1,y)-I(x-1,y))^2 \nonumber \\&& \!\! +(I(x,y+1)-I(x,y-1))^2)^{1/2}
\end{eqnarray}

\begin{equation}\label{Eq.115}
\theta(x,y) = \tan^{-1}(\frac{I(x,y+1)-I(x,y-1)}{I(x+1,y)-I(x-1,y)}).
\end{equation}

\begin{equation}\label{Eq.115-1}
\min \sum_{i=1}^{N}(\| x_{i} - \sum_{j=1}^{M} a_{i}^{(j)} \phi^{(j)} \|^{2} + L)
\end{equation}

\begin{equation}\label{Eq.115-2}
L = \lambda \sum_{j=1}^{M} |a_{i}^{(j)}|.
\end{equation}

Where $x_{i}$ is the SIFT descriptors feature, $a^{j}$ is mostly zero (sparse), $\phi$ is the basis of sparse coding, $\lambda$ is the weights vector.

\section{Location Classification Phase}\label{sec:classifications}
Event classification is a machine learning technique used to predict group membership for data instances. Any classification method uses a set of features or parameters to characterize each object, which called a supervised classification. We prepare  the classes and provide a set of samples. In the training phase, the training set is used to decide how the parameters ought to be weighted and combined in order to separate the various classes of objects. In the testing or application phase, the weights determined in the training set are applied to a set of objects that do not have known classes in order to determine what their classes are likely to be~\cite{Gonzalez1993}. Several major kinds of classification methods like k-nearest neighbor (KNN) classifier and support vector machine (SVM) classifier are used in the proposed system; we briefly   give an overview for each of them in the following subsections.

\subsection{K-Nearest Neighbor (K-NN)}\label{sec:KNN}
The k-nearest-neighbor is a very simple classifier based on the nearest-neighbor approach. In this method, one simply finds in the N-dimensional feature space the closest object from the training set to an object being classified. Since the neighbor is nearby, it is likely to be similar to the object being classified and so is likely to be the same class as that object.

In this classification algorithm, a new object is classified based on majority of K-nearest neighbor category ($K$ is predefined integer), given a query point, the algorithm finds $K$ number of objects or training points closest to the query point. Simply it works based on minimum distance from the searching query to the training one to determine the K-nearest neighbors. After that we gather K-nearest neighbors a simple majority of these K-nearest neighbors is used to be the prediction of the query instance~\cite{Kibriya2007}. We use the Euclidean distance as the distance function as shown in Equation~\eqref{Eq.116}.

\begin{equation}\label{Eq.116}
D(x,y) = \sum_{i=0}^{n} \sqrt{x_{i}^2 - y_{i}^2},
\end{equation}
where $D(x,y)$ is the distance function, $x$ is the query sample, $y$ is the sample from the training set, $n$ is feature dimension. The advantages of K-nearest neighbors are robust to noisy training data especially in inverse square of weighted distance. And disadvantages of K-nearest neighbors are need to determine value of parameter K (number of nearest neighbors), and computation cost is quite high because we need to compute distance of each query instance to all training samples.

\subsection{Artificial Neural Network (ANN)}
The Artificial Neural Network (ANN) is an information processing paradigm that is inspired by the way biological nervous systems, such as the brain, process information. It is composed of a large number of highly interconnected processing elements (neurons) working in unison to solve specific problems. The functions of a biological neuron are modeled by computing a differentiable nonlinear function (such as a sigmoid) for each artificial neuron \cite{cho2002}.

\subsection{Support Vector Machine (SVM)}\label{sec:SVM}
The support vector machine (SVM) algorithm seeks to maximize the margin around a hyperplane that separates a positive class from a negative class. Given a training dataset with n samples $(x_1, y_1), (x_2, y_2), \dots, (x_n, y_n)$, where $x_i$ is a feature vector in a v-dimensional feature space and with labels $y_i \in \{-1,1\}$ belonging to either of two linearly separable classes $C_1$ and $C_2$. Geometrically, the SVM modeling algorithm finds an optimal hyperplane with the maximal margin to separate two classes, which requires to solve the optimization problem, as shown in Equations ~\eqref{Eq.117} and ~\eqref{Eq.118} \cite{Wu2006}.

\begin{equation}\label{Eq.117}
maximize \sum_{i=1}^n \alpha_i - \frac{1}{2} \sum_{i,j=1}^{n}
\alpha_i \alpha_j y_i y_j .K(x_i,x_j),
\end{equation}

\begin{equation}\label{Eq.118}
Subject-to:  \sum_{i=1}^n \alpha_i y_i, 0 \leq \alpha_i \leq C,
\end{equation}

where $\alpha_i$ is the weight assigned to the training sample $x_i$. If $\alpha_i > 0$, $x_i$ is called a support vector. $C$ is a regulation parameter used to trade-off the training accuracy and the model complexity so that a superior generalization capability can be achieved. $K$ is a kernel function, which is used to measure the similarity between two samples. Different choices of kernel functions have been proposed and extensively used in the past and the most popular are the gaussian radial basis function (RBF), polynomial of a given degree, and multi layer perception. These kernels are in general used, independently of the problem, for both discrete and continuous data.

\section{Hajj and Umrah Rituals Classifications}
There are six rite locations during Hajj and Umrah are Tawaf, Sa'y between Safa and Marwa, Standing on mount Arafat, Staying overnight in Muzdalifah, Staying overnight in Mina, and Threw Jamarat. The models defined for this study are described below:\\

\begin{enumerate}

\item{\textbf{Tawaf:}}

Tawaf is one of the Islamic rituals of pilgrimage. During the Hajj and Umrah, Muslims are to circumambulate the Kaaba (most sacred site in Islam) seven times, in a counterclockwise direction, see Fig.~\ref{fig:Tawaf}.\\

\begin{figure}[h]
\begin{center}$
\begin{array}{cc}
\includegraphics[width=2.5in, height=1.7in]{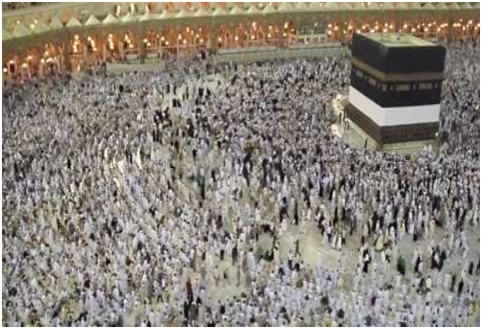}
\end{array}$
\end{center}
\caption{Examples of Tawaf around the Holy Kaaba}
\label{fig:Tawaf}
\end{figure}

\item{\textbf{Sa'y:}}
Pilgrims, whether they are performing Hajj or Umrah perform sa'y after tawaf. Sa'y means endeavoring or making effort. For Hajj, this is held to commemorate Hagar's running between Safa and Marwa seven times in order to find water for her son, Ishmael, whom she was still breast-feeding, see Fig.~\ref{fig:Safa and Marwa}.\\

\begin{figure}
\begin{center}$
\begin{array}{cc}
\includegraphics[width=2.5in, height=1.7in]{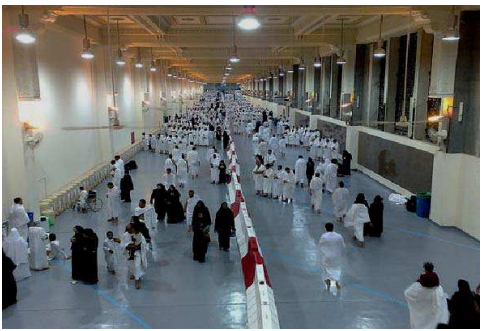}
\end{array}$
\end{center}
\caption{Examples of Sa'y}
\label{fig:Safa and Marwa}
\end{figure}

\item{\textbf{Arafat:}}
The plain of Arafat and Mount Arafat in Saudi Arabia, about three million pilgrims congregated to perform the most important rite of Hajj, or the pilgrimage. This site is significant because it is on the Mount of Mercy that the Prophet Muhammad gave his final sermon. Many pilgrims climb the hill and try to touch the pillar that marks this place. After Arafat, pilgrims will move to Muzdalifah to complete the remaining rites of the pilgrimage, see Fig.~\ref{fig:Arafat}.\\

\begin{figure}
\begin{center}$
\begin{array}{cc}
\includegraphics[width=2.5in, height=1.7in]{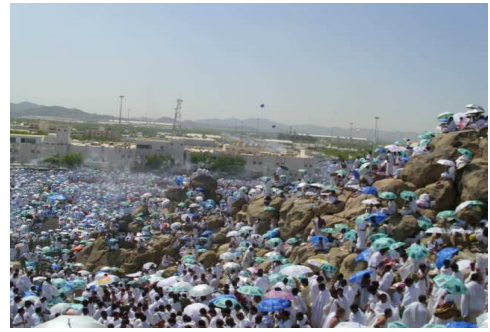}
\end{array}$
\end{center}
\caption{Examples of Arafat}
\label{fig:Arafat}
\end{figure}

\item{\textbf{Muzdalifah:}}
Staying in Muzdalifah is obligatory upon the one performing the Hajj to spend the tenth (10th) of Dhul-Hijjah until the time of Fajr prayer, see Fig.~\ref{fig:Muzdalifah}.\\

\begin{figure}
\begin{center}
\includegraphics[width=7cm, height=5cm]{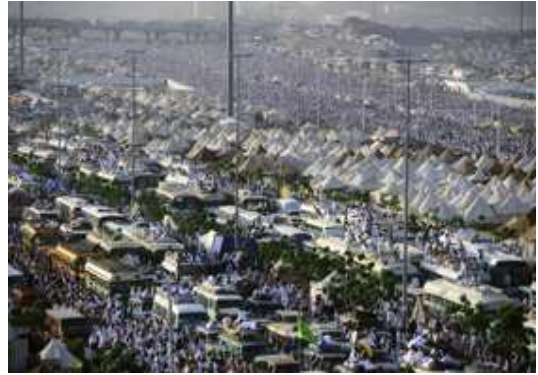}
\caption{Examples of Muzdalifah}
\label{fig:Muzdalifah}
\end{center}
\end{figure}

\item{\textbf{Mina:}}
Mina, seven kilometres east of the Masjid El-Harram is where Hajj pilgrims sleep overnight on the 8th, 11th, 12th (and some even on the 13th) of Dhul Hijjah. It contains the Jamarat, the three stone pillars which are pelted by pilgrims as part of the rituals of Hajj, see Fig.~\ref{fig:Mina}.\\

\begin{figure}[h]
\begin{center}
\includegraphics[width=7cm, height=5cm]{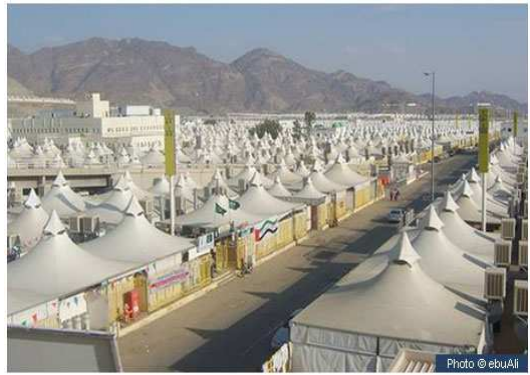}
\label{fig:Mina}\caption{Examples of Mina}\end{center}
\end{figure}

\item{\textbf{Jamarat:}}
The pilgrim who throws Jamarat before Zawal on the eleventh day and the following days will have to throw the pebbles again after Zawal if the days of throwing the pebbles have not yet expired, see Fig.~\ref{fig:Jamarat}.\\

\begin{figure}
\begin{center}
\includegraphics[width=7cm, height=5cm]{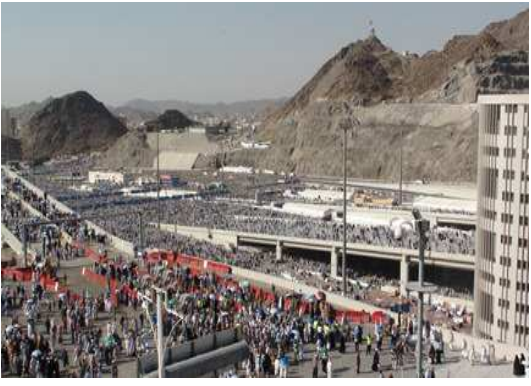}
\caption{Examples of the Jamarat}\label{fig:Jamarat}
\end{center}
\end{figure}

\end{enumerate}

\section{Experimental Results and Analysis}\label{sec:analysis}

The proposed system is evaluated using many videos from Hajj and Umrah datasets~\cite{huer2012}. There are two categories for the quality of the used recorded videos:
\begin{enumerate}[(a)]
\item {\it High resolution}, where frame size of 1280 x 720 pixels.
\item {\it Low resolution}, where frame size of 640 x 480 pixels.
\end{enumerate}
During experiments, both quality categories are covered. All Hajj and Umarh videos are in Audio Video Interleave (AVI) format with a frame rate of $30$ fps.

\begin{figure}[t]
\begin{center}$
\begin{array}{cc}
\includegraphics[width=3.3in, height=2.2in]{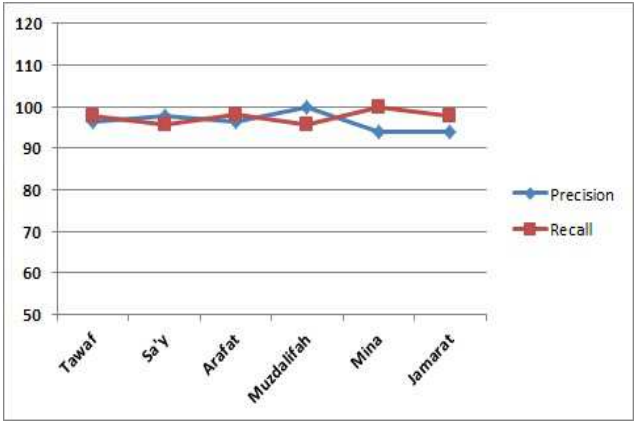}
\end{array}$
\end{center}
\caption{Shot detection results as explained in Table I.}
\label{fig:shotdetection_results}
\end{figure}

Two indicators; namely, \textbf{\textit{recall}} and \textbf{\textit{precision}}, have been designed and calculated for each one of the proposed system phases in order to evaluate the performance of the proposed system and measure the resulted accuracy at each phase. Recall and precision ratios are the basic measures used in evaluating search strategies. Recall is the ratio of the number of relevant records retrieved to the total number of relevant records in the database. Precision is defined as the ratio of the number of relevant records retrieved to the total number of irrelevant and relevant records retrieved. Both recall and precision are usually expressed as a percentage~\cite{Ye2010}. Equations ~\eqref{Eq.119} and ~\eqref{Eq.120} describe calculations of recall and precision ratios, respectively.

\begin{equation}\label{Eq.119}
Recall=\frac{t_p}{t_p +f_n},
\end{equation}

\begin{equation}\label{Eq.120}
Precision=\frac{t_p}{t_p +f_p},
\end{equation}

where, $t_p$, $f_p$, and $f_n$ represent the true positive , false positive , and false negative samples, respectively.\\

The results of the shot boundary detection algorithm are shown in Table \ref{table:table1}. For the collection of Hajj and Umarh videos, the algorithm has correctly identified 281 shot boundaries, while 7 shot boundaries were missed, and in 11 cases a false shot was identified. Accordingly, the average recall and precision were 97.53\% and 96.42\% respectively as shown in Fig.~\ref{fig:shotdetection_results}.

\begin{table*}[t]
\caption{Results of shot detection algorithm for $292$ total experimental images: Among $86$ images, there are $83$ Tawaf correct images,  $2$ Tawaf, which are not discovered, and $3$ images from other classes, which are assigned to Tawaf.}
\label{table:table1}
\begin{center}
\begin{tabular}{|c|c|c|c|c|c|c|c|}
\hline  \hline & &  & && &\\ Event  & Tawaf & Sa'y & Arafat & Muzdalifah & Mina & Jamarat \\
 \hline \hline
 Total & 86 & 45 & 57 & 23 & 32 & 49 \\ \hline
Correct &	83 & 44	& 55 & 23 & 30 & 46 \\ \hline
False & 3	& 1	& 2	& 0	& 2	& 3 \\ \hline
Miss  & 2 & 2	& 1	& 1	& 0	& 1 \\ \hline
Precision & $96.5\%$ & $97.8\%$ & $96.5\%$ & $100\%$ & $93.8\%$ & $93.9\%$ \\ \hline
Recall & $97.6\%$ & $95.7\%$ & $98.2\%$	& $95.8\%$ & $100\%$ & $97.9\%$ \\
  \hline
\end{tabular}
\end{center}
\end{table*}

\begin{figure}[t]
\begin{center}$
\begin{array}{cc}
\includegraphics[width=3.3in, height=2.2in]{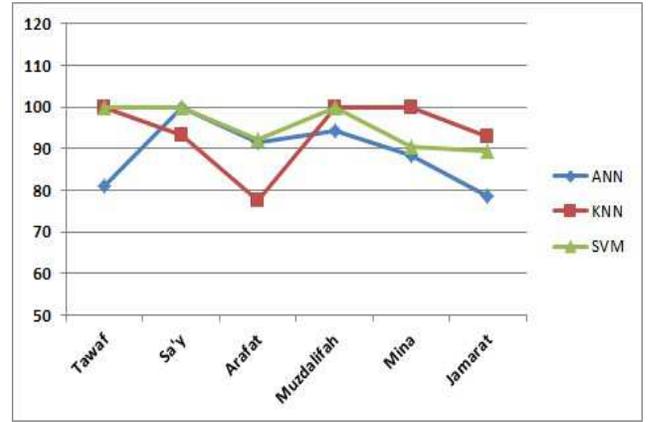}
\end{array}$
\end{center}
\caption{Rital classification results as explained in Table II.}
\label{fig:EventClassficiation_results}
\end{figure}

Table \ref{table:table2} illustrates results of event rites classification using KNN-based, ANN-based, and SVM-based. Compared to the performance results obtained using SVM classifier for the proposed system attained a good result higher than the other classifiers concerning the accuracy which is 95.33\% as shown in Fig.~\ref{fig:EventClassficiation_results}.

\section{Background \& Related Work}\label{sec:previouswork}

Our script-based annotation of human actions is similar in spirit to several recent papers using textual information for automatic image collection from the web \cite{Li2007, Schroff2007} and automatic naming of characters in images \cite{Berg2004} and videos \cite{Everingham2006}. Differently to this work we use more sophisticated text classification tools to overcome action variability in text.

Discriminative interest points can be selected using boosting~\cite{Ke2005, Smith2005}. Given a large number of training samples, boosting can select the most discriminative and representative interest points from a set of randomly generated cuboids. However, boosting usually requires a huge amount of training samples, making it less applicable to small motion datasets.\\

Efros \textit{et. al.}~\cite{Efros2003} proposed an approach to recognize human actions at low resolutions which consisted of a motion descriptor based on smoothed and aggregated optical flow measurements over a spatio-temporal volume centered on a moving figure. This spatial arrangement of blurred channels of optical flow vectors is treated as a template to be matched via a spatio-temporal cross correlation against a database of labeled example actions.

Bobick \textit{et. al.}~\cite{Bobick2001} computed Hu moments of motion energy images and motion-history images to create action templates based on a set of training examples which were represented by the mean and covariance matrix of the moments. Recognition was performed using the Mahalanobis distance between the moment description of the input and each of the known actions.\\

Shechtman and Irani~\cite{Shechtman2005} avoid explicit flow computations by employing a rank-based constraint directly on the intensity information of spatio-temporal cuboids to enforce consistency between a template and a target. Given one example of an action, spatio-temporal patches are correlated against a testing video sequence. Detections are considered to be those locations in space-time which produce the most motion consistent alignments.

Similar to ours, several recent methods explore bag-of-features representations for action recognition \cite{Jhuang2007,Wong2007}, but only address human actions in controlled and simplified settings. Recognition and localization of actions in movies has been recently addressed in~\cite{Laptev2007} for a limited dataset, for example, the manual annotation of two action classes.\\

Nelson~\cite{Rodriguez2008}, developed methods for recognizing human motions by obtaining spatio-temporal templates of motion and periodicity features from a set of optical flow frames. These templates were then used to match the test samples
with the reference motion templates of known activities.

\begin{table*}[t]
\caption{Results of the proposed  rites recognition algorithm: Among $31$ total testing images taken from Mina, on average, the ANN recognizes $27$ of Mina images,  the KNN recognizes all testing samples, and SVM recognizes $28$ of Mina images.}
\label{table:table2}
\begin{center}
\begin{tabular}{|c|c|c|c|c|c|c|}
\hline \hline   & &  & && \\ Event  & Num. of training samples & Num. of testing samples & ANN & KNN & SVM\\
 \hline \hline
Tawaf      & 30 & 30 & $80.95\%$ & $100\%$ & $100\%$ \\ \hline
Sa'y       & 30 & 28 & $100\%$   & $93.33\%$ & $100\%$ \\ \hline
Arafat     & 30 & 25 & $91.30\%$ & $77.42\%$ & $92\%$ \\ \hline
Muzdalifah & 30 & 16 & $94.11\%$ & $100\%$ & $100\%$ \\ \hline
Mina       & 30 & 31 & $88.23\%$ & $100\%$ & $90.5\%$ \\ \hline
Jamarat    & 30 & 19 & $78.57\%$ & $92.86\%$ & $89.5\%$ \\
  \hline
\end{tabular}
\end{center}
\end{table*}
\section{Conclusion}\label{sec:conclusion}
In this paper, we proposed a new event location-based classification system during Hajj and Umrah for video and image scenes. We applied the SIFT and sparse coding algorithms for local features to keep the relationship between these local features. After that we used event recognition classifiers to get the location from the six rituals (Tawaf, Sa'y, Arafat, Muzdalifah, Mina, and Jamarat). We used three type of classifiers (ANN, KNN, and SVM), and apparently SVM classifier achieves the best results.

In our future work, as the Hajj and Umarh (HUER) dataset is limited, so, we will extent it with additional images and videos to check their results with our new system. In addition, we will  apply other algorithms for feature extraction, like (SURF and FAST), and compare the new results  with SIFT method. Certainly, we can achieve more improvement in the artificial neural networks by using feed backward, and Radial Basis Functions.

\section*{Acknowledgments}

This work is funded by a  grant number 11-nan1707-10 from the Long-Term National Plan for Science, Technology and Innovation (LT-NPSTI), the King Abdulaziz City for Science and Technology (KACST), Kingdom of Saudi Arabia. We thank the Science and Technology Unit at Umm A-Qura University for their continued logistics support.

\bigskip
\bigskip

\bigskip
\bigskip
\textit{The consequence of this work is that, we will be able to establish an automatic system to classify any image or video taken from any of the six Hajj and Umrah rituals. With this large data (millions of images and videos) collected annually during Hajj and Umrah seasons either by media or pilgrims, the proposed system will be able to classify, search and extract new information and detect certain incidents that may occur. In addition, researchers from various disciplines (transportation, environmental sciences, medicine,  geography, etc.) can use this computerized system to investigate any research phenomena.}

\bigskip
\bigskip

\bigskip
\textit{Any opinions, findings, and conclusions or recommendations expressed in this work are those of the authors and do not necessarily reflect the views of the Umm Al-Qura University. We thank God for his unlimited support and mercy.  }
\bigskip
\bigskip

\bibliographystyle{plain}

\bibliographystyle{ieeetr}

\end{document}